\documentclass[10pt,twocolumn,letterpaper]{article}

\usepackage[pagenumbers]{iccv} 

\definecolor{iccvblue}{rgb}{0.21,0.49,0.74}
\usepackage[pagebackref,breaklinks,colorlinks,allcolors=iccvblue]{hyperref}
\usepackage{graphicx}
\usepackage{amsmath}
\usepackage{amssymb}
\usepackage{colortbl}
\usepackage{graphicx}
\usepackage{xcolor}
\usepackage{multirow}
\usepackage{booktabs} 
\usepackage{amssymb}
\usepackage{pifont}
\usepackage{algorithm}
\usepackage{algpseudocode}

\usepackage[dvipsnames]{xcolor}
\usepackage{makecell}
\usepackage{rotating}

\newcommand{\rot}[1]{\rotatebox[origin=lb]{60}{\smash{#1}}}

\newcolumntype{x}[1]{>{\centering\arraybackslash}p{#1pt}}
\newcolumntype{y}[1]{>{\raggedright\arraybackslash}p{#1pt}}
\newcolumntype{z}[1]{>{\raggedleft\arraybackslash}p{#1pt}}

\definecolor{mygray}{gray}{0.8}

\def\confName{ICCV}
\def\confYear{2025}

\title{Beyond CLIP Generalization: Against Forward\&Backward Forgetting Adapter for Continual Learning of Vision-Language Models}

\author{Songlin Dong\textsuperscript{\rm 1 \#}{\thanks{Songlin Dong is the corresponding author; \# Songlin Dong and Chenhao Ding are co-first authors}}, Chenhao Ding\textsuperscript{\rm 2 \#}, Jiangyang Li\textsuperscript{\rm 1 }, Jizhou Han\textsuperscript{\rm 1}, Qiang Wang\textsuperscript{\rm 2},Yuhang He\textsuperscript{\rm 1 }, Yihong Gong\textsuperscript{\rm 1}\\
\textsuperscript{\rm 1}College of Artificial Intelligence, Xi’an Jiaotong University\\
\textsuperscript{\rm 2}School of Software Engineering, Xi'an Jiaotong University\\
{\tt\small \{dsl972731417, dch225739\}@stu.xjtu.edu.cn}  
{{\tt\small \{ygong\}@mail.xjtu.edu.cn}}}

\begin{document}
\maketitle
\begin{abstract}
This study aims to address the problem of multi-domain task incremental learning~(MTIL), which requires that vision-language models~(VLMs) continuously acquire new knowledge while maintaining their inherent zero-shot recognition capability. 
Existing paradigms delegate the testing of unseen-domain samples to the original CLIP, which only prevents the degradation of the model's zero-shot capability but fails to enhance the generalization of the VLM further. 
To this end, we propose a novel MTIL framework, named AFA, which comprises two core modules: (1) an against forward-forgetting adapter that learns task-invariant information for each dataset in the incremental tasks to enhance the zero-shot recognition ability of VLMs; (2) an against backward-forgetting adapter that that strengthens the few-shot learning capability of VLMs while supporting incremental learning. Extensive experiments demonstrate that the AFA method significantly outperforms existing state-of-the-art approaches, especially in few-shot MTIL tasks, and surpasses the inherent zero-shot performance of CLIP in terms of transferability. The code is provided in the Supplementary Material.

\end{abstract}    
\begin{figure}
  \centering
    \includegraphics[width=0.98\linewidth]{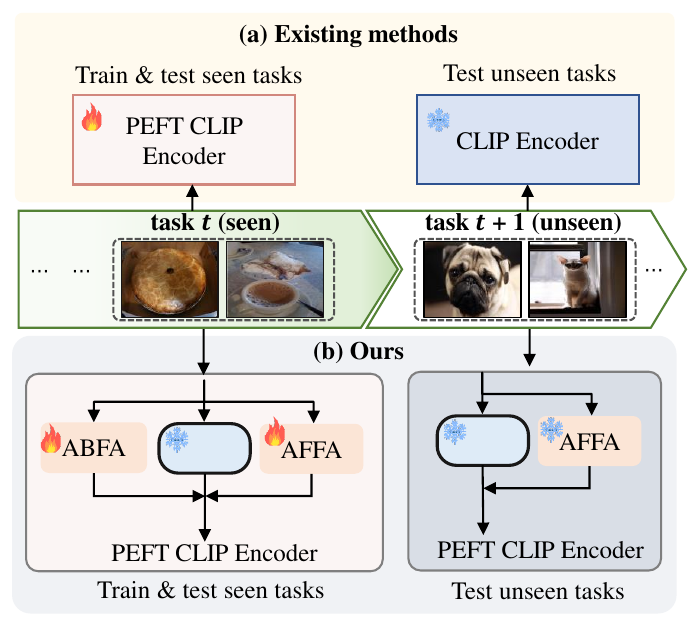}
    \vspace{-0.3cm}
    \caption{Comparison between existing work and our approach:(a) The existing MTIL paradigm.
(b) Our proposed paradigm.}
    \label{fig:head}
  \vspace{-0.3cm}
\end{figure}

\section{Introduction}
\label{sec:intro}


Incremental Learning~\cite{cl_survey,cl_survey2,cl_survey3}~(IL) is a paradigm that requires models to acquire new information while effectively retaining previously learned knowledge. Traditional IL methods typically focus on a single type of increment, such as class-IL or domain-IL, which limits their applicability in complex real-world scenarios. In recent years, the emergence of vision-language models (VLMs), such as CLIP~\cite{clip}, has provided new solutions to these challenges but has also introduced additional difficulties. Specifically, while traditional IL approaches primarily aim to prevent models from forgetting previously learned knowledge~(\textit{i.e} backward forgetting), VLMs face a distinct form of catastrophic forgetting. During incremental learning, they tend to forget the knowledge acquired in the pre-training phase, leading to the degradation of their zero-shot capability~(\textit{i.e} forward forgetting). Consequently, continuously learning new knowledge in VLMs while preserving its zero-shot recognition ability is referred to as Multi-domain Task IL (MTIL)~\cite{zscl}, also known as Domain-Class IL~\cite{zscl-prompt}. This paper aims to address this challenging task.


ZSCL~\cite{zscl} first mitigates knowledge degradation in the continual learning process by combining knowledge distillation from a large-scale reference dataset with parameter regularization. However, this approach relies heavily on computational resources and external data, rendering it impractical in real-world scenarios. Recent methods, such as DIKI~\cite{zscl-prompt} and BCL~\cite{zscl-moe}, have integrated parameter-efficient fine-tuning~(PEFT) techniques (\textit{e.g.}, LoRA~\cite{hu2022lora} or Prompt~\cite{prompt}) with dynamic expansion strategies~\cite{l2p,cada,dytox,gao2023dkt}, thereby introducing a novel paradigm. In particular, these approaches \cite{zscl-moe,zscl-prompt} employ PEFT techniques during the training phase to learn a set of task-specific parameters for each dataset (or domain). 
During testing, distinct strategies are employed to differentiate between seen and unseen datasets: Samples from seen domains are evaluated using task-specific parameters, whereas those from unseen domains are processed by the original CLIP model. 





Although this paradigm alleviates the degradation of VLMs' zero-shot capabilities to some extent, it faces a critical limitation. Specifically, when addressing forward forgetting, these approaches typically assign samples from unseen datasets directly to the original CLIP model during testing without effectively leveraging the knowledge accumulated through continual learning in MTIL tasks~(As shown in Fig.1 (a)). Consequently, while this strategy reduces zero-shot performance degradation, it fails to further enhance the generalization ability of VLMs. Therefore, we pose the following question:


\noindent\textit{\textbf{Can task-invariant knowledge be learned during the MTIL process to enhance model generalization?}}


This problem motivates the core idea of our work: Unlike the existing paradigm of directly assigning unseen-domain samples to the original CLIP model, we propose a novel module—against forward forgetting adapter (AFFA)—which integrates task-invariant knowledge from each task into the CLIP model during incremental learning, thereby enhancing the generalization capability of VLMs~(As shown in Fig.~\ref{fig:head} (b)). Specifically, AFFA employs an efficient fine-tuning structure (e.g., LoRA) that is independent of the incremental tasks, with parameters shared throughout the entire MTIL task. In each MTIL task, the AFFA module fine-tunes VLMs on the current task dataset using contrastive learning, enabling the model to more effectively differentiate between various categories in the feature space and achieve more precise recognition when encountering unseen data.

Another proposed module is the against backward-forgetting adapter (ABFA), which not only enables incremental learning but also emphasizes enhancing the VLM's performance in few-shot scenarios. 
Specifically, we construct a dynamically expanding architecture on a frozen CLIP model based on a multiple-head LoRA MoE structure~\cite{sp-moe1,hydralora,sp-moe2}, where the multi-head LoRA functions as a shared expert, and a progressively expanding task-specific router selects the corresponding expert. This multi-head LoRA structure deploys multiple low-rank (B) matrices for each expert, effectively preventing interference among implicit components within a task and enabling each expert to capture richer task-specific details~\cite{hydralora}, thereby improving parameter efficiency and the model's ability to learn from limited data. In addition, we propose a parameter-free domain distribution selector that, during inference, precisely directs test samples to either the task-specific router within the ABFA framework or the AFFA module for recognition. These two modules constitute the proposed MTIL framework—\textbf{A}gainst Forward $\&$ Backward \textbf{F}orgetting \textbf{A}dapter~(\textbf{AFA}). Extensive experiments demonstrate that our method significantly outperforms state-of-the-art methods, especially in few-shot MTIL tasks, and surpasses the inherent zero-shot performance of CLIP in terms of transfer capability.

The key contributions of this work are as follows: we identify the limitations of existing MTIL approaches in preventing zero-shot capability degradation and propose the AFFA module to enhance the generalization ability of VLMs, enabling their zero-shot recognition performance to surpass that of the original CLIP. Second, for few-shot incremental scenarios, we design the ABFA model to enhance the few-shot learning capability while enabling incremental learning. Extensive experiments on MTIL, few-shot MTIL, CIL, and DIL tasks demonstrate the effectiveness and robustness of our proposed AFA framework.

\section{Related Work}
\label{Related Work}

\noindent\textbf{Vision-Language Models.}
In recent years, exploring semantic correspondence between vision and language in pre-trained VLMs, such as CLIP~\cite{clip}, using large-scale image-text data has been prominent~\cite{align,wang2022vlmixer}.
VLMs align images and texts in a joint embedding space, enabling recognition of nearly unlimited classes. During pre-training, they use contrastive learning, treating paired images and texts as positive samples and different-pair combinations as negative ones. At inference, the text embedding closest to the input image embedding is chosen as the prediction.
Consequently, VLMs can perform zero-shot prediction on unseen tasks and exhibit robust capabilities across a wide range of downstream tasks\cite{coop,gao2024clip,VQA1,VQA3}.

\noindent\textbf{Traditional Incremental Learning.}
Incremental learning (IL) research can be classified into four categories: (1) Regularization methods~\cite{EWC,oewc,MAS,REWC,LWF,lwm} typically impose constraints on the parameter space (e.g., EWC~\cite{EWC}) or the output space (e.g., LWF~\cite{LWF}) to prevent the overwriting of previously acquired knowledge. (2) Rehearsal methods~\cite{ER,der+,ICARL,dong2021few,bic2020,topic,BIC,liu2022model} retain samples associated with previous tasks and replay them when learning new tasks. (3) Architectural-based methods dynamically adjust the model architecture when learning new tasks through techniques such as pruning~\cite{PackNet}, masking~\cite{Piggyback,HAT}, and network expansion~\cite{der2021,foster,dytox,gao2023dkt}, thereby allocating relatively independent training resources for each task. (4) Parameter-tuning methods: Recently, parameter-efficient fine-tuning techniques~\cite{prompt,hu2022lora,adapter1} have been introduced into the IL domain. Their core idea is to optimize pre-trained models for specific tasks by adjusting only a small subset of parameters. For instance, L2P~\cite{l2p} and its variants~\cite{dualprompt,S-prompts,smith2023coda} dynamically generate prompts to guide the adaptation of pre-trained models to new tasks; subsequently, IL methods based on adapter tuning~\cite{cada} and low-rank adaptation~\cite{liang2024inflora} have also emerged.

However, conventional IL settings typically focus solely on either newly introduced classes or changes in domain distribution (i.e., class-IL and domain-IL), which limits their applicability in complex real-world scenarios. Moreover, these methods often struggle to differentiate unseen data and lack zero-shot transfer capability.

\noindent \textbf{Multi-domain Task Incremental Learning.} ZSCL~\cite{zscl} leverages VLMs to incorporate zero-shot transfer capability into continual learning, defining this task as Multi-domain Task IL (MTIL). Initially, task info was required at the testing phase for MTIL tasks~\cite{zscl}. Subsequent refinements~\cite{zscl-moe, zscl-prompt} eliminated the need for task IDs, making the protocol more challenging and realistic. This paper focuses on the scenario where task IDs are \textbf{unavailable} at test time.

Methodologically, ZSCL integrates the zero-shot generalization capacity of the frozen CLIP model via knowledge distillation and uses parameter regularization to prevent knowledge degradation. However, this often brings high computational costs and has poor long-term memory retention. Recent methods combine parameter-efficient fine-tuning~\cite {prompt,hu2022lora,adapter1} and dynamic expansion~\cite{dytox,l2p,cada} strategies to propose VLMs with robust memory and zero-shot transfer. For example, DIKI~\cite{zscl-prompt} employs a fully residual mechanism to integrate new knowledge into the frozen backbone with a distribution-aware integration calibration scheme, minimizing interference and achieving parameter efficiency and good performance. BCL~\cite{zscl-moe} employs classical mixture-of-experts adapters to expand the model for new tasks, with a distribution discriminative auto-selector for input routing to reduce computation.

Compared with existing MTIL approaches~\cite{zscl,zscl-moe,zscl-prompt}, our method not only effectively prevents the degradation of zero-shot generalization but also enhances VLMs' generalization. Moreover, our incremental adapter design, novel to prior works~\cite{liang2024inflora,zscl-moe}, supports incremental learning and boosts model performance in few-shot scenarios.


\begin{figure*}[t]
  \centering
   \includegraphics[width=1\linewidth]{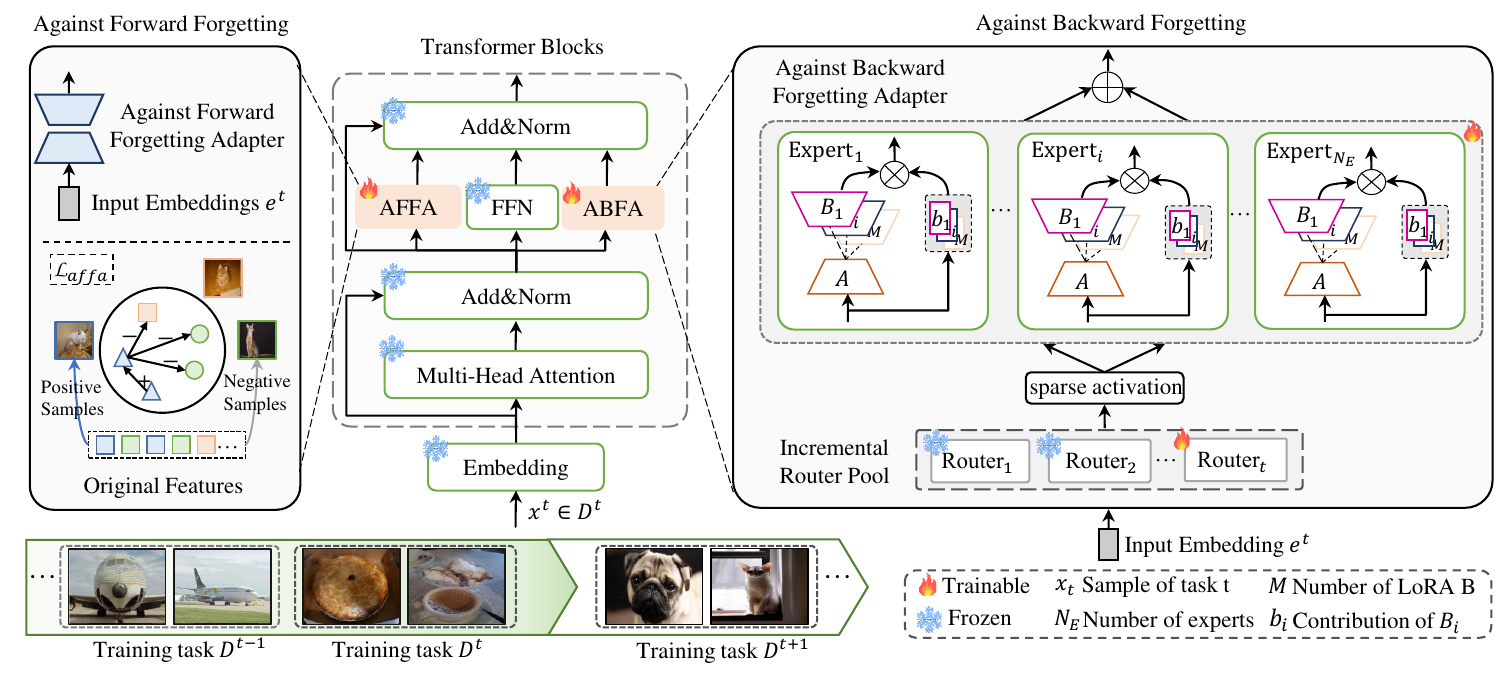}
       \vspace{-0.4cm}
       \caption{\textbf{Overview of AFA method.} (a) AFFA module (left) is trained on \(D^t\) using \(\mathcal{L}_{\text{affa}}\) to boost the generalization of VLMs. (b) the AFBA module (right) expands a task-specific router $h^t$ at stage \(t\) to train \(D^t\) while freezing \(h^1,\ldots,h^{t-1}\) to prevent catastrophic forgetting. In addition, the expert component comprises multi-head LoRA, designed to enhance the model's few-shot learning ability. }
     \vspace{-0.15cm}
   \label{fig:NC-GCD}

\end{figure*}

\section{Method}
\label{sec:method}
\subsection{Problem Definition}
MTIL task can be defined as follows: Given a sequence of \(T\) tasks \(\{\mathbf{T}^t\}_{t=1}^T\) arriving sequentially, where each task $\mathbf{T}^t = \{\mathcal{D}^t, \mathcal{C}^t\}$
is associated with a \textbf{task-specific dataset} $\mathcal{D}^t = \{\mathcal{X}^t, \mathcal{Y}^t\},$
with \(\mathcal{X}^t\) denoting the set of input images and \(\mathcal{Y}^t\) representing the label set. Moreover, the set of category names is defined as $\mathcal{C}^t = \{c^t_j\}_{j=1}^{N_t},$ which maps label indices to the category names used by the VLM (where \(N\) denotes the number of images). The MTIL task requires the VLMs to recognize data from all previously learned tasks \(\mathbf{T}^1,\ldots, \mathbf{T}^t\) during continual learning while also maintaining its generalization ability for the unseen tasks \(\mathbf{T}^{t+1},\ldots, \mathbf{T}^T\) (task IDs remain unknown).

\subsection{Overview Framework}



This paper proposes a novel MTIL framework that enhances zero-shot ability while ensuring robust retention of historical knowledge. Our method is built upon the CLIP~\cite{clip} model, which employs a dual-encoder architecture consisting of an \textit{image encoder} $z(\cdot)$ and a \textit{text encoder} $g(\cdot)$ to extract image and text features from input images and text, respectively.

The framework of the against forward $\&$ backward forgetting adapter~(AFA) is shown in Fig.~\ref{fig:NC-GCD}. The AFA method consists of two adapter modules: against forward-forgetting adapter (AFFA) and against backward-forgetting adapter (ABFA). As shown on the left side of Fig.~\ref{fig:NC-GCD}, the AFFA module employs a LoRA structure that is independent of the incremental tasks, with parameters shared throughout the entire MTIL task. In each MTIL task, the AFFA module fine-tunes VLMs on the current task dataset using contrastive learning to enhance CLIP's zero-shot transfer ability.
As depicted on the right side of Fig.~\ref{fig:NC-GCD}, the ABFA module is built on a dynamic expanding architecture with a multiple-head LoRA mixture of experts (MoE) structure. In MTIL tasks, the multi-head LoRA serves as a shared expert, and a progressively expanding task-specific router selects corresponding experts for incremental learning of VLMs. The multi-head LoRA, deploying several low-rank (B) matrices, enhances few-shot learning by reducing interference among intra-task implicit components.
To eliminate dependency on task IDs during inference, we design a domain distribution selector~(DDS) that, during inference, precisely directs known distribution test samples that are routed to the corresponding router within ABFA, whereas those falling outside are directed to AFFA, thus enabling zero-shot recognition. Next, we will introduce these two modules separately, as well as the training and inference of the AFA method.


\subsection{Against Forward Forgetting Adapter}

To overcome the limitations of existing paradigms~\cite{zscl-moe,zscl-prompt}, we aim to integrate task-invariant knowledge from all tasks into the CLIP model, thereby enhancing its zero-shot transfer capability. As shown in Fig.~\ref{fig:NC-GCD} (left), we propose a LoRA structure shared across \( T \) tasks and embed it into the feedforward network (FFN) of the Transformer block to train \( T \) datasets.  
For the embedding \( e^t \) of the \( t \)-th task, its forward propagation process can be expressed as:
\begin{equation}
 y^t_{affa}=BAe^{t}.
\end{equation}

Recent studies~\cite{chen2020simple,li2024enhancing,clip} have shown that contrastive learning enables the model to better differentiate between different classes in the feature space, thus facilitating more accurate recognition when encountering unseen data. Therefore, we fine-tune CLIP on the \( T \) tasks through contrastive learning. For the visual feature \( \mathbf{v}_{i} \) and the text feature \( \mathbf{w}_{j} \), we first compute the similarity between the visual and text features as follows:
\begin{equation}
    l_{i,j} = \frac{\mathbf{v}_i \cdot \mathbf{w}_j}{\|\mathbf{v}_i\| \|\mathbf{w}_j\|}.
\end{equation}

The contrastive learning formula for visual-text pairs is as follows:
\begin{equation}
    \mathcal{L}_{\text{text-image}} = -\sum_{i=1}^{N} \frac{1}{|P(i)|} \sum_{j \in P(i)} \log \frac{\exp(l_{i,j})}{\sum_{k=1}^{N} \exp(l_{i,k})},
\end{equation}
where \( P(i) \) is the set of sample indices that share the same category label \( y_i \) as \( i \).
Similarly, we can obtain the text-visual contrastive loss as follows:
\begin{equation}
    \mathcal{L}_{\text{image-text}} = -\sum_{i=1}^{N} \frac{1}{|P(i)|} \sum_{j \in P(i)} \log \frac{\exp(l_{j,i})}{\sum_{k=1}^{N} \exp(l_{k,i})}.
\end{equation}

Finally, the loss function for training the AFFA module is formulated as follows:
\begin{equation}
    \label{eq:affa}
    {\mathcal{L}_{affa}} = \mathcal{L}_{\text{text-image}} + \mathcal{L}_{\text{image-text}}.
\end{equation}








\begin{table*}[t]
\setlength\tabcolsep{5pt}
\centering
\setlength{\belowcaptionskip}{1mm}

\label{tab:finalacc}
{
\fontsize{8pt}{8pt}\selectfont
\resizebox{0.9\textwidth}{!}{
\begin{tabular}{y{70}x{25}|x{25}*{10}{x{17}}|x{22}}
\toprule
 \textbf{Method }& \rot{Extra data} & \rot{Aircraft} & \rot{Caltech101} & \rot{CIFAR100} & \rot{DTD} & \rot{EuroSAT} & \rot{Flowers} & \rot{Food} & \rot{MNIST} & \rot{OxfordPet} & \rot{Cars} & \rot{SUN397} & \rot{Average} \\ \midrule

\quad Zero-shot & & 24.3 & 88.4 & 68.2 & 44.6 & 54.9 & 71.0 & 88.5 & 59.4 & 89.0 & 64.7 & 65.2 & 69.4${}^{*}$ \\
\quad FFT & & 62.0 & 96.2 & 89.6 & 79.5 & 98.9 & 97.5 & 92.7 & 99.6 & 94.7 & 89.6 & 81.8 & 89.3 \\ \midrule

\textbf{Transfer} \\
\quad LwF~\cite{LWF} & $\checkmark$ & & 74.5 & 56.9 & 39.1 & \textbf{51.1} & 52.6 & 72.8 & 60.6 & 75.1 & 30.3 & 55.9 & 56.9 \\
\quad LwF-VR~\cite{lwm} & $\checkmark$ & & 77.1 & 61.0 & 40.5 & 45.3 & 54.4 & 74.6 & 47.9 & 76.7 & 36.3 & 58.6 & 57.2 \\
\quad WiSE-FT~\cite{wise} & $\checkmark$ & & 73.5 & 55.6 & 35.6 & 41.5 & 47.0 & 68.3 & 53.9 & 69.3 & 26.8 & 51.9 & 52.3  \\

\quad L2P~\cite{l2p} & $\times$ & & 65.6 & 50.9 & 30.4 & 41.4 & 49.3 & 71.8 & 36.3 & 77.5 & 55.3 & 53.4 & 53.2 \\
\quad Dual-prompt~\cite{dualprompt} & $\times$ & & 56.7 & 51.4 & 28.7 & 33.7 & 45.6 & 70.9 & 59.5 & 77.7 & 49.5 & 50.4 & 52.4 \\
\quad S-Prompts~\cite{S-prompts} & $\times$ & & 67.3 & 49.4 & 26.4 & 39.7 & 47.1 & 70.2 & 34.3 & 78.9 & 56.7 & 52.2 & 52.2 \\ \midrule


\quad ZSCL~\cite{zscl} & $\checkmark$ & & 86.0 & 67.4 & {45.4} & 50.4 & {69.1} & {87.6} & 61.8 & 86.8 & 60.1 & {66.8} & 68.1 \\

\quad DIKI~\cite{zscl-prompt} & $\times$ & & 88.4 & \textbf{69.0} & 43.2 & 48.2 & 67.4 & 85.2 & {63.0} & {87.9} & {63.8} & 66.2 & {68.3} \\ 

\quad BCL~\cite{zscl-moe} & $\checkmark$ & & 87.9 & 68.2 & 44.4 & 49.9 & 70.7 & 88.7 & 59.7 & 89.1 & 64.5 & 65.5 & 68.9 \\ 

\rowcolor{mygray} \quad AFA~(Ours) & $\times$ & & \textbf{88.6} & 67.9 & \textbf{45.7} & \textbf{54.8} &	\textbf{71.3} & \textbf{88.5} & \textbf{64.4} & \textbf{89.7} &	\textbf{64.7} & \textbf{67.0} & \textbf{70.3} \\

\midrule

\textbf{Avg.} \\
\quad LwF~\cite{LWF} & $\checkmark$ & 36.3 & 86.9 & 72.0 & 59.0 & 73.7 & 60.0 & 73.6 & 74.8 & 80.0 & 37.3 & 58.1 & 64.7 \\
\quad LwF-VR~\cite{lwm} & $\checkmark$ & 29.6 & 87.7 & 74.4 & 59.5 & 72.4 & 63.6 & 77.0 & 66.7 & 81.2 & 43.7 & 60.7 & 65.1 \\
\quad WiSE-FT~\cite{wise} & $\checkmark$ & 26.7 & 86.5 & 64.3 & 57.1 & 65.7 & 58.7 & 71.1 & 70.5 & 75.8 & 36.9 & 54.6 & 60.7 \\

\quad L2P~\cite{l2p} & $\times$ & 38.0 & 85.2 & 78.2 & 61.3 & 72.9 & 74.9 & 79.7 & 59.1 & 82.0 & 59.7 & 55.4 & 67.9 \\
\quad Dual-prompt~\cite{dualprompt} & $\times$ & 37.8 & 84.3 & 78.6 & 60.1 & 71.1 & 73.2 & 79.1 & 73.9 & 82.3 & 55.1 & 52.8 & 68.0 \\
\quad S-Prompts~\cite{S-prompts} & $\times$ & 37.5 & 92.5 & 77.5 & 58.2 & 76.4 & 74.1 & 78.8 & 57.9 & 83.0 & 60.8 & 54.4 & 68.3 \\ \midrule

\quad ZSCL~\cite{zscl} & $\checkmark$ & 45.1 & 92.0 & 80.1 & 64.3 & 79.5 & 81.6 & \textbf{89.6} & 75.2 & 88.9 & 64.7 & {68.0} & 75.4 \\

 \quad DIKI~\cite{zscl-prompt} & $\times$ & 45.1 & \textbf{95.5} & {83.1} & {64.8} & {79.9} & {83.5} & 87.0 & {76.2} & {89.6} & {67.0} & 67.1 & {76.3} \\ 

 \quad BCL~\cite{zscl-moe} & $\checkmark$ & 50.2 & 91.9 & 83.1 & 69.4 & 78.9 & 84.0 & 89.1 & 73.7 & 89.3 & 67.7 & 66.9 & 76.7 \\  

 \rowcolor{mygray} \quad AFA~(Ours) & $\times$ & \textbf{56.0} & 93.7 & \textbf{85.3} & \textbf{69.5} & \textbf{80.0} & \textbf{85.4} &	89.0 & \textbf{76.8} & \textbf{91.0} & \textbf{67.9} &	\textbf{68.6} & \textbf{78.5}  \\
\midrule

\textbf{Last} \\
\quad LwF~\cite{LWF} & $\checkmark$ & 26.3 & 87.5 & 71.9 & 66.6 & 79.9 & 66.9 & 83.8 & \textbf{99.6} & 92.1 & 66.1 & 80.4 & 74.6 \\
\quad LwF-VR~\cite{lwm} & $\checkmark$ & 20.5 & 89.8 & 72.3 & 67.6 & 85.5 & 73.8 & 85.7 & \textbf{99.6} & 93.1 & 73.3 & 80.9 & 76.6 \\
\quad WiSE-FT~\cite{wise} & $\checkmark$ & 27.2 & 90.8 & 68.0 & 68.9 & 86.9 & 74.0 & 87.6 & \textbf{99.6} & 92.6 & 77.8 & 81.3 & 77.7 \\
\quad L2P~\cite{l2p} & $\times$ & 38.0 & 87.1 & 84.2 & 72.9 & 86.0 & 96.1 & 89.2 & 99.0 & 94.1 & 79.6 & 76.0 & 82.0 \\
\quad Dual-prompt~\cite{dualprompt} & $\times$ & 37.8 & 87.1 & 84.6 & 71.8 & 89.2 & 96.3 & 89.1 & 99.1 & {94.5} & 79.9 & 76.5 & 82.3 \\
\quad S-Prompts~\cite{S-prompts} & $\times$ & 37.5 & 95.1 & 83.7 & 70.2 & 97.5 & 96.5 & 89.0 & 99.1 & 94.0 & 79.5 & 75.8 & 83.4 \\
\midrule

\quad ZSCL~\cite{zscl} & $\checkmark$ & 40.6 & 92.2 & 81.3 & 70.5 & 94.8 & 90.5 & \textbf{91.9} & 98.7 & 93.9 & 85.3 & 80.2 & 83.6 \\

\quad DIKI & $\times$ & 45.2 & \textbf{95.7} & {86.3} & {72.9} & \textbf{98.0} & {97.0} & 89.2 & 99.4 & 94.2 & 81.6 & 76.6 & {85.1}\\

 \quad BCL~\cite{zscl-moe} & $\checkmark$ & 49.8 & 92.2 & 86.1 & 78.1 & 95.7 & 94.3 & 89.5 & 98.1 & 89.9 & 81.6 & 80.0 & 85.0 \\  

\rowcolor{mygray} \quad AFA~(Ours) & $\times$ & \textbf{56.0} & 94.2 & \textbf{89.2} & \textbf{78.4} & 94.4 & \textbf{97.1} &	89.5 & 98.7 & \textbf{94.6} & 82.1 & \textbf{84.7} & \textbf{87.2} \\

\bottomrule
\end{tabular}}
}
\caption{Comparison with state-of-the-art methods on MTIL benchmark in \textit{Transfer}, \textit{Avg.}, and \textit{Last} scores (\%). Metric ``transfer'' represents the model zero-shot ability retention after being trained on each task. To ensure a fair comparison, we exclude Aircraft when evaluating frozen CLIP's transfer capability.
All compared methods employ the CLIP model with the ViT-B/16 backbone~\cite{vit}.}
\label{tab:full_mtil}
\vspace{-0.25cm}
\end{table*}
\subsection{Against Backward Forgetting Adapter}\label{sec:abfa}

ABFA enables the CLIP model to learn a sequence of task-specific datasets continuously. To achieve this, we employ a mixture-of-experts (MoE) adapter to construct a scalable incremental learning paradigm. Specifically, the MoE consists of a set of shared experts \(\{E_i\}_{i=1}^{N_E}\) and task-specific routers that expand as tasks progress, where \(N_E\) denotes the predefined number of experts. We integrate this MoE into each Transformer block's feedforward network~(FFN).

Given an embedding \( e^{t} \) of task \( t \), a set of contribution weights \( w^{t} \) is first calculated for different experts using the task-specific router \( h^t(\cdot) \):
\begin{equation}
\label{eq:router}
      w^{t}=Router(e^{t})={\rm{softmax}}(h^t(e^{t})),
\end{equation}
where \( h(\cdot) \) is a linear layer without bias. Based on the weights \( w^{t} = \{ w^{t}_{i} \}_{i=1}^{N_{E}} \), the output of the MoE adapter composed of \( N_{E} \) experts can be expressed as follows:
\begin{equation}
    \label{eq:moe}
    y^{t}=\sum_{i=1}^{N_{E}}w^t_{i}\cdot f_{i}(e^{t}),
\end{equation}
where \( f_{i}(\cdot) \) represents the \( i \)-th expert.  
To mitigate interference between incremental tasks, we introduce a sparse selection strategy~\cite{sp-moe1,sp-moe2} to regulate expert outputs. Specifically, based on the weights \( w^{t} \), we activate the top \( k \) experts with the highest contributions. Consequently, Eq. \eqref{eq:router} can be reformulated as:
\begin{equation}
    \label{eq:srouter}
    \hat{w}^{t}={\rm{Top_{k}}}({\rm
    {softmax}}(h^t(e^{t})),k).
\end{equation}

Research shows that classical LoRA suffers from interference among implicit task components, which limits its ability to capture task-specific details~\cite{hydralora}. This, in turn, leads to suboptimal performance in low-data scenarios. To address it, we employ multi-head LoRA as experts, incorporating \( M \) low-rank matrices (\( B \) matrices) within a single expert. This approach enhances both parameter efficiency and the model’s ability to learn from limited data. It can be formulated as:
\begin{equation}
    \label{eq:srouter22}
    f_{i}=\sum_{i=1}^{M}b_{i}\cdot B_{i}Ae^{t}, 
\end{equation}
where \( b_i \) follows the same formula as \( w \) in Eq.~\eqref{eq:router}. Thus, the final output of the AFFA module is obtained as follows:
\begin{align}
    \label{eq:moe22}
    y^{t}_{abfa}=\sum_{j=1}^{k}\hat{w}^t_{j}\cdot \sum_{i=1}^{M}b_{i}\cdot B_{i} Ae^{t}.
\end{align}



\begin{table*}[t]
\setlength\tabcolsep{5pt}
\centering
{
\footnotesize
\fontsize{8pt}{8pt}\selectfont
\resizebox{0.9\textwidth}{!}{
\begin{tabular}{y{70}x{25}|x{25}*{10}{x{17}}|x{22}}
\toprule
\textbf{ Method} & \rot{Extra data} & \rot{Aircraft} & \rot{Caltech101} & \rot{CIFAR100} & \rot{DTD} & \rot{EuroSAT} & \rot{Flowers} & \rot{Food} & \rot{MNIST} & \rot{OxfordPet} & \rot{Cars} & \rot{SUN397} & \rot{Average} \\ \midrule

\quad Zero-shot & & 24.3 & 88.4 & 68.2 & 44.6 & 54.9 & 71.0 & 88.5 & 59.4 & 89.0 & 64.7 & 65.2 & 69.4${}^{*}$ \\
\quad 5-shot FFT & & 30.6 & 93.5 & 76.8 & 65.1 & 91.7 & 92.9 & 83.3 & 96.6 & 84.9 & 65.4 & 71.3 & 77.5 \\ \midrule

\textbf{{Transfer}} \\
\quad LwF~\cite{LWF} & $\checkmark$ & & 72.1 & 49.2 & 35.9 & 44.5 & 41.1 & 66.6 & 50.5 & 69.0 & 19.0 & 51.7 & 50.0 \\
\quad LwF-VR~\cite{lwm} & $\checkmark$ & & 82.2 & 62.5 & 40.1 & 40.1 & 56.3 & 80.0 & 60.9 & 77.6 & 40.5 & 60.8 & 60.1 \\
\quad WiSE-FT~\cite{wise} & $\checkmark$ & & 77.6 & 60.0 & 41.3 & 39.4 & 53.0 & 76.6 & 58.1 & 75.5 & 37.3 & 58.2 & 57.7  \\

\midrule


\quad ZSCL~\cite{zscl} & $\checkmark$ & & {84.0} & {68.1} & {44.8} & {46.8} & {63.6} & {84.9} & {61.4} & {81.4} & {55.5} & {62.2} & {65.3} \\

\quad DIKI~\cite{zscl-prompt} & $\times$ & & 88.3 & \textbf{68.4} & 43.8 & 47.8 & 71.1 & 85.8 & 59.4 & 89.2 & \textbf{65.8} & 62.6 & 68.2 \\ 

\quad BCL~\cite{zscl-moe} & $\checkmark$ & & {87.9} & {68.2} & {44.1} & {48.1} & {64.7} & \textbf{{88.8}} & \textbf{{69.0}} & {89.1} & {64.5} & {65.1} & 68.9 \\ 

\rowcolor{mygray} \quad AFA~(Ours) & $\times$ & & \textbf{88.5} & 67.9 & \textbf{45.7} &\textbf{54.7} & \textbf{71.2} & 88.6 & 63.7 & \textbf{89.7} & 64.7 & \textbf{66.8} & \textbf{70.2} \\

\midrule

\textbf{{Avg.}} \\
\quad LwF~\cite{LWF} & $\checkmark$ & 23.5 &77.4  &43.5 &41.7 &43.5  &52.2 &54.6 & 63.4 & 68.0& 21.3& 52.6&49.2 \\
\quad LwF-VR~\cite{lwm} & $\checkmark$ &24.9 &{89.1} &64.2 &53.4 &54.3  &70.8 &79.2 &66.5  &79.2 & 44.1& 61.6&62.5\\
\quad WiSE-FT~\cite{wise} & $\checkmark$ & {32.0} & 87.7 & 61.0 &{55.8} & {68.1} & 69.3&76.8 &{71.5}  &77.6 &42.0 &59.3&63.7 \\

\midrule

\quad ZSCL~\cite{zscl} & $\checkmark$ & 28.2& 88.6 &{66.5} & 53.5&56.3  &{73.4} &{83.1} & 56.4 & {82.4} & {57.5}&{62.9}&{64.4} \\

 \quad DIKI~\cite{zscl-prompt} & $\times$ & 25.1 & 	88.5 & 	69.0 & 	43.8 & 	48.2 & 	72.8 & 	85.8 & 	58.8 	& 89.1 & \textbf{67.2} & 	63.3 & 	64.7 \\ 

 \quad BCL~\cite{zscl-moe} & $\checkmark$ & 30.0 & {89.6} & \textbf{{73.9}} & {58.7}& {69.3} &{79.3} & {88.1}& \textbf{{76.5}} & {89.1}& {65.3}&{65.8} & {71.4} \\  

 \rowcolor{mygray} \quad AFA~(Ours) & $\times$ &\textbf{44.7} & \textbf{91.9} & 73.8 &\textbf{ 58.9} & \textbf{76.9} & 	\textbf{83.4} & \textbf{88.2 }& 73.8 & 	\textbf{89.7} & {66.0} & \textbf{67.5} & \textbf{74.1}
  \\
\midrule

\textbf{{Last}} \\
\quad LwF~\cite{LWF} & $\checkmark$ & 22.1 & 58.2 & 17.9 & 32.1 & 28.1 & 66.7 & 46.0 & 84.3 & 64.1 & 31.5 & 60.1 & 46.5 \\
\quad LwF-VR~\cite{lwm} & $\checkmark$ & 22.9 & 89.8 & 59.3 & 57.1 & 57.6 & 79.2 & 78.3 & 77.7 & 83.6 & 60.1 & 69.8 & 66.9 \\
\quad WiSE-FT~\cite{wise} & $\checkmark$ & {30.8} & 88.9 & 59.6 & {60.3} & {80.9} & 81.7 & 77.1 & {94.9} & 83.2 & 62.8 & 70.0 & {71.9} \\
\midrule

\quad ZSCL~\cite{zscl} & $\checkmark$ & 26.8 & 88.5 & {63.7} & 55.7 & 60.2 & {82.1} & {82.6} & 58.6 & {85.9} & {66.7} & {70.4} & 67.4 \\

\quad DIKI & $\times$ & 23.7 &	89.6 &	68.7 &	43.6 &	51.4 &	68.7 &	85.3 &	53.5 &	88.2 &	\textbf{73.5} &	70.3 &	65.1  \\

 \quad BCL~\cite{zscl-moe} & $\checkmark$ & {30.1} & {89.3} &{74.9}  &\textbf{{64.0}} & {82.3} &  {89.4}&{87.1} & 89.0 & {89.1} &{69.5} & {{72.5}}  & {76.1} \\  

\rowcolor{mygray} \quad AFA~(Ours) & $\times$ & \textbf{44.7}  & 	\textbf{92.3}  & \textbf{75.1}  & 63.9  & \textbf{89.6}  & 	\textbf{93.5}  & \textbf{87.8}  & \textbf{91.5}  & \textbf{89.5}  & {71.7}  & \textbf{73.7} &  \textbf{79.4} \\ 

\bottomrule
\end{tabular}}
}
\caption{Comparison with state-of-the-art methods on few-shot MTIL benchmark in \textit{Transfer}, \textit{Avg.}, and \textit{Last} scores (\%). Metric ``transfer'' represents the model zero-shot ability retention after being trained on each task. To ensure fair comparison, we exclude Aircraft when evaluating frozen CLIP's transfer capability. All compared methods employ the CLIP model with the ViT-B/16 backbone~\cite{vit}. 
}
\label{tab:mtil_few}
\vspace{-2mm}
\end{table*}

\subsection{Training and Inference of AFA}

\noindent\textbf{Domain distribution selector.} For task \( \mathbf{T}^t \), we extract features using a frozen CLIP model and apply K-Means clustering to obtain \( K \) task prototypes \( P^{t} = \{ p^{t}_{1}, \dots, p^{t}_{K} \} \).  
For any test sample \( x\), we compute its task score based on the similarity with all learned task prototypes \( P = \{ P^{1}, \dots, P^{t} \} \): $s^{t}=\frac{1}{K}{<s,P^{t}>}$, where \( <,> \) denotes cosine similarity. The task domain of \( x\) is determined by the task \( t \) with the highest score \( \max(s_{t=1}^t) \).  To mitigate interference with zero-shot capabilities, samples with scores below a predefined threshold are classified as ``unseen” task domains.  

\noindent\textbf{Train and Inference.} (1) Training Phase: For task \( \mathbf{T}^t \), the ABFA module extends a new router $h^t$ and, in conjunction with the shared expert set, trains task task \( \mathbf{T}^t \) using a cross-entropy (CE) loss. After training, the router $h^t$ is frozen to preserve its output for the $t$-th task and prevent catastrophic forgetting. Subsequently, the task-shared AFFA module is trained with a contrastive loss (see Eq.~\eqref{eq:affa}) to enhance the model's zero-shot capabilities. It is important to note that during the training of any module, both the original CLIP and the other module remain frozen to avoid mutual interference. (2) Inference Phase: First, a distribution score \( s \) is computed by the domain distribution selector. If \( s \) is below a predefined threshold, the test sample is classified as belonging to an "unseen domain" and is processed by the AFFA module; otherwise, the router corresponding to the task with the highest score in the ABFA module is selected for testing.


\section{Experiments}
\label{sec:experiment}
\begin{table*}[!t]
    \centering
    \small
    \resizebox{0.9\linewidth}{!}{
    \begin{tabular}{lc|cc|cc|cc|cc|cc|cc}
        \toprule 
        \multirow{4}{*}{\makecell[l]{Method}} & \multirow{4}{*}{\makecell[l]{Backbone}} & \multicolumn{6}{c|}{CIFAR100} & \multicolumn{6}{c}{Tiny-ImageNet} \\
        \cmidrule(lr){3-8} \cmidrule(lr){9-14}
         &  & \multicolumn{2}{c}{10 step} & \multicolumn{2}{c}{20 step} & \multicolumn{2}{c|}{50 step} & \multicolumn{2}{c}{5 step} & \multicolumn{2}{c}{10 step} & \multicolumn{2}{c}{20 step} \\
        \cline{3-8} \cline{9-14}
         &  &  Avg. & Last & Avg. & Last & Avg. & Last &  Avg. & Last & Avg. & Last & Avg. & Last \\
        \midrule
        CLIP Zero-shot & \multirow{5}{*}{CLIP} & 74.47 & 65.92 & 75.20 & 65.74 & 75.67 & 65.94 & 69.62 & 65.30 & 69.55 & 65.59 & 69.49 & 65.30 \\
        Fine-tune &  & 65.46 & 53.23 & 59.69 & 43.13 & 39.23 & 18.89 & 61.54 & 46.66 & 57.05 & 41.54 & 54.62 & 44.55 \\
        LwF \cite{LWF} &  & 65.86 & 48.04 & 60.64 & 40.56 & 47.69 & 32.90 & 60.97 & 48.77 & 57.60 & 44.00 & 54.79 & 42.26 \\
        iCaRL  \cite{ICARL} & & 79.35 & 70.97 & 73.32 & 64.55 & 71.28 & 59.07 & 77.02 & 70.39 & 73.48 & 65.97 & 69.65 & 64.68 \\
        LwF-VR \cite{lwm} &  & 78.81 & 70.75 & 74.54 & 63.54 & 71.02 & 59.45 & 77.56 & 70.89 & 74.12 & 67.05 & 69.94 & 63.89 \\ \midrule
        ZSCL \cite{zscl} & \multirow{3}{*}{CLIP} & 82.15 & 73.65 & 80.39 & 69.58 & 79.92 & 67.36 & 80.27 & 73.57 & 78.61 & 71.62 & 77.18 & 68.30 \\
        BCL~\cite{zscl-moe} &  & 85.21 & 77.52 & 83.72 & 76.20 & 83.60 & 75.24 & 81.12 & 76.81 & 80.23 & 76.35 & 79.96 & 75.77 \\
        \rowcolor{mygray} AFA~(Ours) &  & \textbf{86.74} & \textbf{79.33} & \textbf{85.39} & \textbf{77.97} & \textbf{84.33} & \textbf{76.36} & \textbf{82.14} & \textbf{78.37} & \textbf{81.36} & \textbf{78.01} & \textbf{81.09} & \textbf{77.54} \\
        \bottomrule
    \end{tabular}}
    \vspace{-0.05cm}
    \caption{Comparison of state-of-the-art methods on CIFAR100 and TinyImageNet datasets in class-incremental learning settings.}
    \vspace{-0.2cm}
    \label{tab:cil_combined}
\end{table*}

\begin{table}[h]
  \centering
  \footnotesize
  \renewcommand{\arraystretch}{1}
  \resizebox{1\linewidth}{!}{
  \begin{tabular}{lcccc}
    \hline
    \multirow{2}{*}{Method} & \multirow{2}{*}{Backbone} & \multirow{2}{*}{Buffer size} & \multicolumn{2}{c}{Avg. Accuracy} \\
    \cline{4-5}
    & & & DomainNet & CORe50 \\
    \midrule
    DyTox~\cite{dytox} & ViT-B/16 & 50/class & 62.94 & 79.21 \\
    \midrule
    LwF~\cite{LWF} & \multirow{4}{*}{ViT-B/16} & \multirow{4}{*}{0/class} & 49.19 & 75.45 \\
    L2P~\cite{l2p} & & & 40.12 & 78.33 \\
    Dual-prompt~\cite{dualprompt} & & & 43.79 & 80.25 \\
    S-Prompts~\cite{S-prompts} & & & 50.62 & 83.13 \\
    \midrule
    S-Prompts~\cite{S-prompts} & \multirow{4}{*}{CLIP} & \multirow{4}{*}{0/class} & 67.78 & 89.06 \\
    PINA~\cite{wang2024non} & & & 69.06 & 87.38 \\
    BCL~\cite{zscl-moe} & & & 66.47 & 87.61 \\
    \rowcolor{mygray} AFA (Ours) & & & 70.22 & 90.81 \\
    \hline
  \end{tabular}}
  \vspace{-0.05cm}
  \caption{Comparison results on the DomainNet and CORe50. The domain ID of the test image is unknown in the DIL setting.}
  \vspace{-0.3cm}
  \label{tab:domain}
\end{table}

\vspace{-2pt}
\subsection{Experimental Setup}
\label{sec:setup}
\vspace{-2pt}
\textbf{Dataset and Benchmarks.} This work evaluates the proposed method on four tasks: Multi-domain Task-IL~(MTIL), Few-shot MTIL~(MTIL-FS), Class-IL~(CIL), and Domain-IL~(DIL).
(1) MTIL: We follow the two-order training protocol proposed in ZSCL~\cite{zscl}. The MTIL task comprises 11 different datasets that are learned sequentially.
The main paper applies the Order-I protocol, and Order-II is provided in Appendix Table~\textcolor{Maroon}{8}. (2) MTIL-FS: Based on the MTIL training Order-I protocol, we adopt only 5 samples per class for each dataset to simulate a data-scarce scenario. (3) CIL: Evaluations are done by splitting classes within one dataset ~\cite{lucir}. CIFAR100 (100 classes) is divided into tasks of 10, 20, or 50 subsets. For TinyImageNet (200 classes), the first stage trains on 100 classes, and the remaining 100 are split into tasks of 5, 10, or 20 subsets to test adaptability to class distributions. (4) DIL: Following the protocol in~\cite{S-prompts}, we evaluate two DIL datasets with multiple domains to evaluate adaptability to domain distributions. CORe50 ~\cite{core50} has 11 domains (8 for training, 3 for testing). DomainNet ~\cite{domainnet} is a large-scale dataset with 6 domains that have big differences.

\vspace{-2pt}
\noindent\textbf{Evaluation metrics.} To evaluate our method on the MTIL and MTIL-FS tasks, we use the metrics `Transfer,' `Average,' and `Last' proposed in~\cite{zscl}. Specifically, the `Transfer' metric evaluates the model's zero-shot transfer capability on unseen data, i.e., its ability to counteract forward forgetting. The `Last' metric assesses the model's retention of historical knowledge, i.e., its resistance to backward forgetting.
The `Average' metric, the mean of `Transfer' and `Last',  indicates overall performance. For the CIL and DIL tasks, following ~\cite{lucir, S-prompts},
we compute the average accuracy across all tasks (`Average'). In the CIL setting, we report the accuracy of the final task (`Last'). `Buffer' denotes the number of exemplars cached per class.

\noindent\textbf{Implementation Details.} For fair comparison, we follow the protocol in~\cite{zscl} and employ the CLIP model with the ViT-B/16 backbone~\cite{vit} for all experiments. For structure details, we apply the AFFA and ABFA modules to both the vision and language branches~(Appendix Table~\textcolor{Maroon}{12}). Specifically, ABFA is integrated into all transformer layers, whereas AFFA is incorporated only in the final layer. For ABFA, the total number of experts is $N_{E}$=22~(Fig.~\ref{fig:aba_expert_number} (a)), with each expert consisting of 1 A and 4 B components~(Table.~\ref{tab:aba_ima}). In terms of sparse activation, the number of activated experts $k$=2~(Fig.~\ref{fig:aba_expert_number} (b)). For training details, we use AdamW as the optimizer, training for 1000 iterations in the MTIL setting and 300 iterations in the MTIL-FS setting. More implementation details are shown in the Appendix.

\subsection{Comparison with State-of-The-Arts}

\noindent\textbf{Multi-domain Task Incremental Learning.} Table~\ref{tab:full_mtil} shows a comparison between our method and the current SOTA approaches on the MTIL task. Experimental results indicate that our method, AFA, significantly outperforms the previous best approaches on all three metrics.
(1) Transfer: We observed that traditional IL methods (such as LWF and L2P) generally lack zero-shot transfer capability, resulting in poor transfer scores and evident forward forgetting. In contrast, AFA achieves a transfer score of \textbf{70.3\%}, significantly surpassing MTIL methods such as ZSCL (68.1\%, an improvement of \textbf{2.3\%} ), DIKI (68.3\%), and BCL (68.9\%). Notably, AFA’s Transfer performance even exceeds CLIP’s zero-shot capability (69.4\%, an improvement of \textbf{0.9\%}), which demonstrates that AFA enhances the model’s generalization ability.
(2) Last: For the ``last" metric, AFA achieves a score of \textbf{87.2\%}, showing significant improvements over DIKI (85.1\%) and BCL (85.0\%). This suggests that AFA effectively mitigates forgetting throughout the learning process, demonstrating superior resistance to backward forgetting.
(3) Average: AFA attains an average accuracy of \textbf{78.5\%}, which is \textbf{2.2\%} and \textbf{1.8\%} higher than DIKI (76.3\%) and BCL (76.7\%), respectively.


\noindent\textbf{Few-shot MTIL.} We conducted experiments on the 5-shot MTIL-FS benchmark in Table~\ref{tab:mtil_few}. The experimental findings are consistent with those observed in the MTIL task: our method, AFA, achieved the best performance across all three metrics, with particularly significant improvements in anti-forgetting capabilities (as measured by `Last' metrics). Specifically, in terms of Transfer ability, AFA markedly enhanced the model’s zero-shot performance in few-shot scenarios, attaining a score of \textbf{70.2\%}, which represents an improvement of \textbf{0.8\%} over the frozen CLIP’s zero-shot performance, and up to \textbf{2.0\%} higher than state-of-the-art methods such as BCL and DIKI, as well as \textbf{4.9\%} higher than the ZSCL. Regarding the last accuracy metric, AFA demonstrated even more pronounced performance advantages in the MTIL-FS task, achieving a \textbf{3.3\%} improvement over the second-best approach, BCL. These results provide compelling evidence of the effectiveness of AFA in few-shot scenarios.

\noindent\textbf{Class Incremental Learning.} We conducted CIL experiments to assess our method's ability to adapt to class distributions. For the CIL task, we assigned two experts per session and used one router. Table~\ref{tab:cil_combined} shows our method's comparative performance on CIFAR100 and TinyImageNet. Results show that our method outperforms traditional IL methods and MTIL approaches, proving the proposed AFA method's effectiveness and scalability in handling single-domain CIL challenges.

\noindent\textbf{Domain Incremental Learning.} 
We evaluated AFA's ability to adapt to domain distributions via DIL experiments. Table~\ref{tab:domain} compares our method's performance on DomainNet and CORe50 datasets. Results show that our method outperforms DIL and MTIL approaches,  achieving up to \textbf{1.75\%} over the best DIL method and up to \textbf{3.75\%} over the MTIL method, demonstrating its effectiveness and scalability in single-domain DIL tasks.


\begin{table}[!h]
    \centering
    \small
    \resizebox{0.98\linewidth}{!}{
    
    \begin{tabular}{lcc|cc|cc}
    \toprule
    Method & Transfer & $\Delta$ & Average & $\Delta$ & Last & $\Delta$  \\
\midrule
BCL~\cite{zscl-moe} & 68.9&  \textcolor{MidnightBlue}{-0.5} & 76.7 & \textcolor{Maroon}{+ 11.4}& 85.0& \textcolor{Maroon}{+19.7}     \\ \midrule
CLIP Zero-shot & 69.4 & 0.0 & 65.3  & 0.0 & 65.3 & 0.0\\
+Adapter & 45.0 &  \textcolor{MidnightBlue}{-24.4} & 57.0 &  \textcolor{MidnightBlue}{-8.3} &71.5 & \textcolor{Maroon}{+6.2} \\        
+ABFA & 69.3 & \textcolor{MidnightBlue}{-0.1}& 77.9 &\textcolor{Maroon}{+12.6}  & 87.2 & \textcolor{Maroon}{+21.9} \\
 +ABFA+AFFA & \textbf{70.3} & \textcolor{Maroon}{\textbf{+0.9}} & \textbf{78.5} &\textcolor{Maroon}{\textbf{+13.2}}  & \textbf{87.2} & \textcolor{Maroon}{\textbf{+21.9}} \\
    \bottomrule
    \end{tabular}}
     \vspace{-0.05cm}
    \captionof{table}{The effectiveness of our proposed modules. The experiment is conducted under the MTIL setting.}
    \vspace{-0.2cm}
    \label{tab:aba_module}
\end{table}

\subsection{Ablation Study}
\noindent\textbf{The Effectiveness of Each Module.} We conducted ablation studies on the MTIL benchmark, as shown in Table~\ref{tab:aba_module}, where `Adapter' denotes fine-tuning CLIP using a single adapter. (1) Transfer Ability: Directly fine-tuning the adapter results in a significant decline in zero-shot performance. The SOTA method only partially alleviates the degradation, with a \textbf{0.5\%} drop, whereas our proposed ABFA module exhibits almost no degradation (only a \textbf{0.1\%} decline) and with the AFFA module, improves generalization by \textbf{0.9\%} over original CLIP, proving AFA's effectiveness in enhancing zero-shot capability. (2) Incremental Ability (`Last'): The ABFA module significantly enhances CLIP’s incremental learning capability, achieving improvements of \textbf{21.9\%} over CLIP and \textbf{2.1\%} over the SOTA method.
(3) `Average': The results show that incorporating the ABFA module improves overall model performance by \textbf{12.6\%}. With the further integration of the AFFA module, the model’s zero-shot ability is further enhanced, ultimately reaching a best average performance of \textbf{78.5\%}, which is a \textbf{13.2\%} improvement.  

\begin{table}[h]
	\centering
   
        \resizebox{0.98\linewidth}{!}{
	\begin{tabular}{l|cc|cc}
		\toprule		
            Method  &MTIL-FS Last.&  $\Delta$ & MTIL  Last &$\Delta$ \\
		\midrule
         CLIP Zero-shot & 65.3 &  0.0 &  65.3 &  0.0\\
        Ours(1A/1B)  & 76.3 & \textcolor{Maroon}{+11.0} &  86.1 &  \textcolor{Maroon}{+20.8}\\
        
		Ours(1A/2B) & 77.8 & \textcolor{Maroon}{+12.5} &  86.6 & \textcolor{Maroon}{+21.3}  \\
        
		Ours(1A/4B) & 79.4 & \textcolor{Maroon}{\textbf{+14.1}}  & 87.2 & \textcolor{Maroon}{\textbf{+21.9}}\\
        
        Ours(1A/6B)& 78.9 & \textcolor{Maroon}{+13.6}  & 86.9 &  \textcolor{Maroon}{+21.6}\\
    
		\bottomrule
	\end{tabular}}
    \vspace{-0.1cm}
	\caption{The effectiveness of multiple-heads structure in ABFA module. ``1A/mB'' indicates that each expert has one low-rank matrix A and M high-rank matrices B, respectively.}
    \vspace{-0.1cm}
	\label{tab:aba_ima}
\end{table}

        
        
        
    

 
\noindent\textbf{The Effectiveness of Multiple-heads Structure.} We conducted ablation studies, as shown in Table~\ref{tab:aba_ima}. We observed that, compared to the original structure (1A/1B), using multiple B structures consistently enhanced the last accuracy across tasks. Notably, the best-performing 1A/4B structure yielded only a \textbf{1.1\%} improvement on the MTIL task but a substantial \textbf{3.1\%} boost on the MTIL-FS task, demonstrating the effectiveness of the multi-head architecture in few-shot learning. Moreover, more pronounced improvements were observed on fine-grained datasets (e.g., Aircraft, DTD) and more complex datasets (e.g., SUN397) (Appendix Table \textcolor{Maroon}{14}), which supports our analysis in the Sec.~\ref{sec:abfa}.

\noindent\textbf{Computational Cost.} In Table~\ref{tab:cost}, we compare our method with other MTIL approaches in terms of computational cost. Firstly, compared to the MTIL method ZSCL, which adopts a full fine-tuning strategy, our method exhibits significant advantages across all three metrics. Secondly, concerning the SOTA method BCL, which is also based on adapters, our approach reduces the training parameters (M), GPU burden (MiB), and iteration time by \textbf{49.65\%}, \textbf{19.21\%}, and \textbf{1.27\%}, respectively. These results strongly indicate that our method outperforms on the MTIL task while significantly reducing training computational costs.
\begin{table}[t]
	\centering
         \resizebox{0.96\linewidth}{!}{
	\begin{tabular}{c >{\centering\arraybackslash}p{2.4cm} >{\centering\arraybackslash}p{2.1cm}>{\centering\arraybackslash}p{1.8cm}}
		\toprule		Method & Train Params $\downarrow$ &GPU $\downarrow$ & Times $\downarrow$ \\
		\midrule
            \textcolor{Black}{LWF}~\cite{LWF}&\textcolor{Black}{149.6M} & \textcolor{Black}{32172MiB} & \textcolor{Black}{1.54s/it}\\
            
		  ZSCL~\cite{zscl} &149.6M & 26290MiB & 3.94s/it \\
        
            \rowcolor{gray!20}
            BCL~\cite{zscl-moe}&	\ \ 59.8M&	22358MiB	& 1.58s/it \\
            
            \rowcolor{mygray} 
            AFA~(Ours) & \ \ 30.1M & 18064MiB & 1.56s/it \\
            
            \textbf{$\Delta$} &	\textbf{\textcolor{Maroon}{-49.65\%}}&	\textbf{\textcolor{Maroon}{-19.21\%}}	& \textbf{\textcolor{Maroon}{-1.27\%}} \\
            
		\bottomrule
	\end{tabular}}
    \vspace{-5pt}
	\caption{\textcolor{Black}{Comparison of computational cost during training on training parameters, GPU burdens, and training times of each iteration. \textbf{$\Delta$} is the improvement to the SOTA method BCL~\cite{zscl-moe}.}}
	\label{tab:cost}
 \vspace{-15pt}
\end{table}

\begin{figure}[h]
\vspace{-0.1cm}
\begin{center}
\centerline{\includegraphics[width=\linewidth]{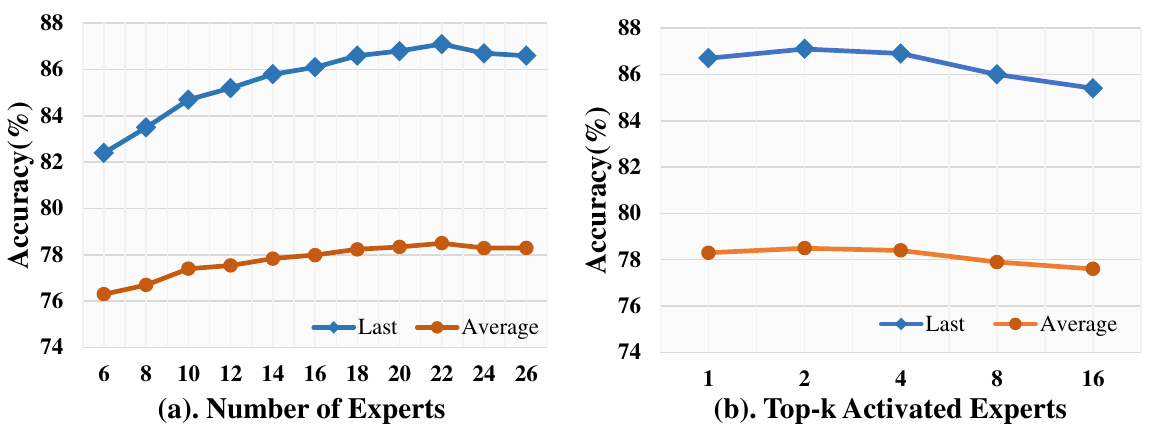}}
 \vspace{-8pt}
\caption{Impact of numbers of experts and top-k activated experts. The experiment
is conducted under the MTIL setting.}
\vspace{-20pt}
\label{fig:aba_expert_number}
\end{center}
\end{figure}

\noindent\textbf{Hyper-parameter Sensitivity Analysis:}
\begin{itemize}
    \item \textbf{Impact of Numbers of Experts.} As shown in Fig.~\ref{fig:aba_expert_number} (a), we analyze the impact of the number of experts on MoE-Adapter performance in an MTIL task with \( T \) task-specific routers. Performance improves as the expert count increases, especially at lower values, but plateaus beyond \( 2T \) (i.e., 22 experts). Thus, we adopt \( 2T \) experts in our experiments.

     \item \textbf{Hyper-parameter $k$.} In the MTIL task, we conducted ablation studies on the number of activated experts \( k \) in the AFFA module. Results show peak performance at \( k=2 \), while excessive activation degrades performance due to task interference. Thus, we set \( k=2 \) in our experiments.
\end{itemize}

\section{Conclusion}
In this paper, we propose the against forward $\&$ backward forgetting adapter to tackle the multi-domain task incremental learning problem in VLMs. First, we propose the AFFA module, which learns task-invariant information from each dataset in the incremental tasks to enhance the zero-shot recognition capability of VLMs. Second, we propose the ABFA module, which not only supports incremental learning but also improves VLM performance in few-shot scenarios. Extensive experiments across four tasks demonstrate the effectiveness and robustness of our approach.

{
    \small
    \bibliographystyle{ieeenat_fullname}
    \bibliography{ICCV2025-Author-Kit/ref}
}
\clearpage
\setcounter{page}{1}
\maketitlesupplementary

\section{Additional Experiment Results}
\label{sec:add_results}

In this section, we first provide a detailed analysis of the test accuracy of our method across various tasks. To further validate its robustness, we also evaluate its performance on ORDER-II.

\noindent\textbf{Detail Test Accuracy.} Table \ref{tab:tab:supp_whole_full_shot} presents the performance of our method on the full-sample MTIL benchmark. Table \ref{tab:supp_whole_few_shot} presents the performance of our method on the few-shot MTIL benchmark.

\noindent\textbf{Order-II Compare Result.} To further verify the robustness of our method, we conduct experiments on the Order-II sequence, with the results shown in Table \ref{tab:supp_mtil_order2}. The Order-II sequence includes StanfordCars, Food, MNIST, OxfordPet, Flowers, SUN 397, Aircraft, Caltech 101, DTD, EuroSAT, and CIFAR 100. As observed, the proposed method outperforms state-of-the-art approaches across all three metrics in both settings. Notably, the zero-shot transfer capability of the proposed method \textbf{surpasses} the upper bound of the pre-trained CLIP model.
\begin{table*}[t]
\setlength\tabcolsep{5pt}
\centering
\setlength{\belowcaptionskip}{1mm}

\label{tab:finalacc}
{
\fontsize{8pt}{9pt}\selectfont
\resizebox{0.95\textwidth}{!}{
\begin{tabular}{y{70}x{25}|x{25}*{10}{x{17}}|x{22}}
\toprule
 \textbf{Method} & \rot{Extra data} & \rot{Cars} & \rot{Food}  & \rot{MNIST} & \rot{OxfordPet} & \rot{Flowers} & \rot{SUN397} & \rot{Aircraft} & \rot{Caltech101} & \rot{DTD} & \rot{EuroSAT} & \rot{CIFAR100} & \rot{Average} \\ 
 \midrule

\quad Zero-shot & & 64.7 & 88.5 & 59.4 & 89.0 & 71.0 & 65.2 & 24.3 & 88.4 & 44.6 & 54.9 & 68.2 & 65.4${}^{*}$ \\
\quad FFT & & 89.6 & 92.7 & 99.6 & 94.7 & 97.5 & 81.8 & 62.0 & 95.1 & 79.5 & 98.9 & 89.6 & 89.2 \\ \midrule

{Transfer} \\
\quad LwF~\cite{LWF} & $\checkmark$ & & 87.8&58.5&71.9&46.6&57.3&12.8&81.4&34.5&34.5&46.8& 53.2 \\
\quad iCaRL~\cite{ICARL} & $\checkmark$ & & 86.1 & 51.8 & 67.6 & 50.4 & 57.9 & 11.0 & 72.3 & 31.2 & 32.7 & 48.1 & 50.9 \\
\quad LwF-VR~\cite{lwm} & $\checkmark$ & &88.2 & 57.0 & 71.4 & 50.0 & 58.0 & 13.0 & 82.0  &34.4  &29.3  &47.6 & 53.1 \\
\quad WiSE-FT~\cite{wise} & $\checkmark$ & & 87.2 & 57.6 & 67.0 & 45.0 & 54.0 & 12.9 & 78.6 & 35.5 & 28.4&  44.3 & 51.1  \\

\midrule


\quad ZSCL~\cite{zscl} & $\checkmark$ & & 88.3 & 57.5 & 84.7 & 68.1 & {64.8} & {21.1} & {88.2} & {45.3} & {55.2} & {68.2} & 64.1 \\

\quad DIKI~\cite{zscl-prompt} & $\times$ & & 85.8 & 59.8 & 89.1 & 71.8&  62.6&  24.3 & 93.3&  42.7 & 46.8 & 67.8 & 64.4 \\ 

\quad BCL~\cite{zscl-moe} & $\checkmark$ & & {88.8} &{59.5} & {89.1} & {69.9} & 64.4 & {18.1} & 86.9 & 43.7 & {54.6} & {68.2} & 64.3\\ 
\rowcolor{mygray} \quad AFA~(Ours) & $\times$ & & 88.9& 64.5& 89.7& 71.4& 66.9& 24.8	& 88.6& 45.6& 55.0& 68.0& \textbf{66.3} \\

\midrule
{Avg.} \\
\quad LwF~\cite{LWF} & $\checkmark$ & 49.0 & 77.0 & 92.1 & 85.9 & 66.5 & 67.2 & 20.9 & 84.7 & 44.6 & 45.5 & 50.5 & 62.2 \\
\quad iCaRL~\cite{ICARL} & $\checkmark$ & 52.0 & 75.9 & 77.4 & 74.6 & 58.4 & 59.3 &  11.7& 79.6 & 42.1 & 43.2 & 51.7 & 56.9 \\
\quad LwF-VR~\cite{lwm} & $\checkmark$ & 44.9 & 75.8 & 91.8  &85.3 & 63.5  &67.6 & 16.9 & 84.9  &44.0  &40.6 & 51.3  &60.6 \\
\quad WiSE-FT~\cite{wise} & $\checkmark$ & 52.6 & 79.3 & 91.9 & 83.9  & 63.4 & 65.2 & 23.3 & 83.7  & 45.4 & 40.0 & 48.2 & 61.5 \\


\midrule

\quad ZSCL~\cite{zscl} & $\checkmark$ & 81.7& {91.3} & 91.1 & {91.0} & 82.9 & {72.5} & {33.6} & {89.7} & {53.3} & {62.8} & {69.9} &   74.5 \\

 \quad DIKI~\cite{zscl-prompt} & $\times$ & 81.9&  88.9&  92.1&  92.8 & 87.7&  70.3 & 34.3 & 94.2&  51.5&  56.1&  69.5&  74.5 \\ 

 \quad BCL~\cite{zscl-moe} & $\checkmark$ & {84.9} & {89.9} & 89.3 & {91.4} & {86.2} & 72.2 & {33.4} & 89.4 & {53.3} & 61.4 & {69.9} & {74.7} \\  

 \rowcolor{mygray} \quad AFA~(Ours) & $\times$ & 85.4& 90.2 &	92.2 &93.4 &	87.7 &	76.5 &	38.7 &	90.8 &	54.3 &	62.6 &	69.9 &	\textbf{76.5} \\
\midrule
{Last} \\
\quad LwF~\cite{LWF} & $\checkmark$ & 34.6 & 69.6 & 99.3 & 88.7 & 61.1 & 72.5 & 32.5 & 88.1 & 65.6 & 90.9 & 87.9 & 71.9 \\
\quad iCaRL~\cite{ICARL} & $\checkmark$ & 35.8 & 93.0 & 77.0 & 70.2 & 83.3 & 88.5 & 90.4 & 86.7 & 93.2 & 81.2 & {81.9} & 80.1 \\
\quad LwF-VR~\cite{lwm} & $\checkmark$ & 20.5 & 89.8 & 72.3 & 67.6 & 85.5 & 73.8 & 85.7 & {99.6} & 93.1 & 73.3 & 80.9 & 76.6 \\
\quad WiSE-FT~\cite{wise} & $\checkmark$ & 35.6 & 76.9 & {99.5} & 89.1 & 62.1 & 71.8 & 27.8 & 90.8 & 67.0 & 85.6 & 87.6 &  72.2 \\
\midrule

\quad ZSCL~\cite{zscl} & $\checkmark$ & 78.2 & {91.1} & 97.6 & {92.5} & 87.4 & {78.2} & {45.0} & 92.3 & 72.7 & {96.2} & 86.3 &  83.4 \\

\quad DIKI & $\times$ & 81.9& 89.2 &99.4 &94.3 &96.8 &76.7 &46.3 &95.9 &74.8 &98.3 &86.6 &85.5\\

 \quad BCL~\cite{zscl-moe} & $\checkmark$ &84.1 & 88.5 & 94.0 & {91.8} & {94.1} & {77.8} & {50.4} & {93.3} & {77.1} & 87.7 & 86.6 & {84.1} \\  

\rowcolor{mygray} \quad AFA~(Ours) & $\times$ & 85.4 &	90.3& 	98.3 &	94.8 &	97.0 	&84.5 	&55.3 	&94.7 	&77.6 	&96.8 &	89.1 &	\textbf{87.6}  \\

\bottomrule
\end{tabular}}
}
\caption{Comparison with state-of-the-art methods on MTIL~(Order-II) benchmark in \textit{Transfer}, \textit{Avg.}, and \textit{Last} scores (\%). Metric ``transfer'' represents the model zero-shot ability retention after being trained on each task. To ensure fair comparison, we exclude Aircraft when evaluating frozen CLIP's transfer capability. 
}
\label{tab:supp_mtil_order2}
\vspace{-2mm}
\end{table*}

\begin{table*}[h]
    \centering
    \resizebox{0.90\linewidth}{!}{
    \begin{tabular}{l>{\centering\arraybackslash}p{1cm} >{\centering\arraybackslash}p{1cm} >{\centering\arraybackslash}p{1cm}>{\centering\arraybackslash}p{1cm} >{\centering\arraybackslash}p{1cm} >{\centering\arraybackslash}p{1cm}>{\centering\arraybackslash}p{1cm} >{\centering\arraybackslash}p{1cm} >{\centering\arraybackslash}p{1cm}>{\centering\arraybackslash}p{1cm} >{\centering\arraybackslash}p{1cm} >{\centering\arraybackslash}p{1cm}}
        \toprule
             & \rotatebox {45}{Aircraft} & \rotatebox {45}{Caltech101} & \rotatebox {45}{CIFAR100} & \rotatebox {45}{DTD}& \rotatebox {45}{EuroSAT} & \rotatebox {45}{Flowers} & \rotatebox {45}{Food} & {\rotatebox {45}{MNIST}} & {\rotatebox {45}{OxfordPet}} & {\rotatebox {45}{Cars}} & {\rotatebox {45}{SUN397}} & \\
            \midrule
            Transfer & & 88.6 & 67.9 & 45.7 & 54.8 & 71.3 & 88.5 & 64.4 & 89.7 & 64.7 & 67.0 & \colorbox{Orchid}{70.3}\\
            \midrule
            Aircraft & \colorbox{YellowGreen}{56.0} & 88.6 & 67.9 & 45.7 & 54.8 &71.3 & 88.5&64.4 &89.7 &64.7 &67.0 & \\
            Caltech101 &56.0 &\colorbox{YellowGreen}{94.2} & 67.9 & 45.7 & 54.8 &71.3 & 88.5&64.4 &89.7 &64.7 &67.0 & \\
            CIFAR100 & 56.0&94.2&\colorbox{YellowGreen}{89.2}&45.7 & 54.8 &71.3 & 88.5&64.4 &89.7 &64.7 &67.0 & \\
            DTD &56.0 &94.2&89.2&\colorbox{YellowGreen}{78.4}&54.8 &71.3 & 88.5&64.4 &89.7 &64.7 &67.0 & \\
            EuroSAT &56.0&94.2&89.2&78.4&\colorbox{YellowGreen}{94.4}&71.3 & 88.5&64.4 &89.7 &64.7 &67.0 & \\
            Flowers &56.0&94.2&89.2&78.4&94.4&\colorbox{YellowGreen}{97.1}&88.5&64.4 &89.7 &64.7 &67.0 & \\
            Food &56.0&94.2&89.2&78.4&94.4&97.1&\colorbox{YellowGreen}{89.5}&64.4 &89.7 &64.7 &67.0 & \\
            MNIST &56.0&94.2&89.2&78.4&94.4&97.1&89.5&\colorbox{YellowGreen}{98.7}&89.7 &64.7 &67.0 & \\
            OxfordPet&56.0&94.2&89.2&78.4&94.4&97.1&89.5&98.7&\colorbox{YellowGreen}{94.3}&64.7 &67.0 & \\
            Cars &56.0&94.2&89.2&78.4&94.4&97.1&89.5&98.7&94.3&\colorbox{YellowGreen}{82.1}&67.0&\\
            SUN397 &56.0&94.2&89.2&78.4&94.4&97.1&89.5&98.7&94.3&82.1&\colorbox{YellowGreen}{84.7}& \colorbox{Dandelion}{87.1} \\
            \midrule
            Average &56.0&93.7&85.3&69.5&80.0&85.4&89.0&76.8&91.0&67.9&68.6& \colorbox{Tan}{78.5}\\
        \bottomrule
    \end{tabular}}
    \caption{Accuracy (\%) of our method (Ours) on the MTIL benchmark. Each row represents the performance on every dataset of the model trained after the corresponding task. \colorbox{Orchid}{Transfer}, \colorbox{Tan}{Average}, and \colorbox{Dandelion}{Last} metrics are shown in color.}
    \label{tab:tab:supp_whole_full_shot}
    
\end{table*}

\begin{table*}[h]
    \centering
    \resizebox{0.90\linewidth}{!}{
    \begin{tabular}{l>{\centering\arraybackslash}p{1cm} >{\centering\arraybackslash}p{1cm} >{\centering\arraybackslash}p{1cm}>{\centering\arraybackslash}p{1cm} >{\centering\arraybackslash}p{1cm} >{\centering\arraybackslash}p{1cm}>{\centering\arraybackslash}p{1cm} >{\centering\arraybackslash}p{1cm} >{\centering\arraybackslash}p{1cm}>{\centering\arraybackslash}p{1cm} >{\centering\arraybackslash}p{1cm} >{\centering\arraybackslash}p{1cm}}
        \toprule
             & \rotatebox {45}{Aircraft} & \rotatebox {45}{Caltech101} & \rotatebox {45}{CIFAR100} & \rotatebox {45}{DTD}& \rotatebox {45}{EuroSAT} & \rotatebox {45}{Flowers} & \rotatebox {45}{Food} & {\rotatebox {45}{MNIST}} & {\rotatebox {45}{OxfordPet}} & {\rotatebox {45}{Cars}} & {\rotatebox {45}{SUN397}} & \\
            \midrule
            Transfer & & 88.5 & 67.9 & 45.7 & 54.7 & 71.2 & 88.6 & 63.7 & 89.7 & 64.7 & 66.8 & \colorbox{Orchid}{70.2}\\
            \midrule
            Aircraft & \colorbox{YellowGreen}{44.7} & 88.5 & 67.9 & 45.7 & 54.7 & 71.2 & 88.6 & 63.7 & 89.7 & 64.7 & 66.8 & \\
            Caltech101 &44.7 &\colorbox{YellowGreen}{92.2} & 67.9 & 45.7 & 54.7 & 71.2 & 88.6 & 63.7 & 89.7 & 64.7 & 66.8 & \\
            CIFAR100 & 44.7&92.2&\colorbox{YellowGreen}{75.1}&45.7 & 54.7 & 71.2 & 88.6 & 63.7 & 89.7 & 64.7 & 66.8 & \\
            DTD &44.7 &92.2&75.1&\colorbox{YellowGreen}{63.9}&54.7 & 71.2 & 88.6 & 63.7 & 89.7 & 64.7 & 66.9 & \\
            EuroSAT &44.7 &92.3&75.1&63.9&\colorbox{YellowGreen}{89.6}&71.2 & 88.6 & 63.8 & 89.7 & 64.7 & 66.9 & \\
            Flowers &44.7 &92.3&75.1&63.9&89.6&\colorbox{YellowGreen}{93.5}&88.6 & 63.8 & 89.7 & 64.7 & 66.9 & \\
            Food &44.7 &92.3&75.1&63.9&89.6&93.5&\colorbox{YellowGreen}{87.8}&63.8 & 89.7 & 64.7 & 66.9 & \\
            MNIST &44.7 &92.3&75.1&63.9&89.6&93.5&87.9&\colorbox{YellowGreen}{91.5}&89.7 &64.7 &66.9 & \\
            OxfordPet&44.7 &92.3&75.1&63.9&89.6&93.5&87.8&91.5&\colorbox{YellowGreen}{89.5}&64.7 &66.9 & \\
            Cars &44.7 &92.3&75.1&63.9&89.6&93.5&87.9&91.5&89.5&\colorbox{YellowGreen}{71.7}&66.9&\\
            SUN397 &44.7 &92.3&75.1&63.9&89.6&93.5&87.8&91.5&89.5&71.7&\colorbox{YellowGreen}{73.7}& \colorbox{Dandelion}{79.4} \\
            \midrule
            Average &44.7&91.4&73.8&58.9&76.9&83.4&88.2&73.8&89.7&66.0&67.5& \colorbox{Tan}{74.0}\\
        \bottomrule
    \end{tabular}}
    \caption{Accuracy (\%) of our method (Ours) on the few-shot MTIL benchmark. Each row represents the performance on every dataset of the model trained after the corresponding task. \colorbox{Orchid}{Transfer}, \colorbox{Tan}{Average}, and \colorbox{Dandelion}{Last} metrics are shown in color.}
    \label{tab:supp_whole_few_shot}
    
\end{table*}

\section{Additional Experimental Setup}
\label{sec:add_setup}
\begin{figure}[t]
\begin{center}
\centerline{\includegraphics[width=\linewidth]{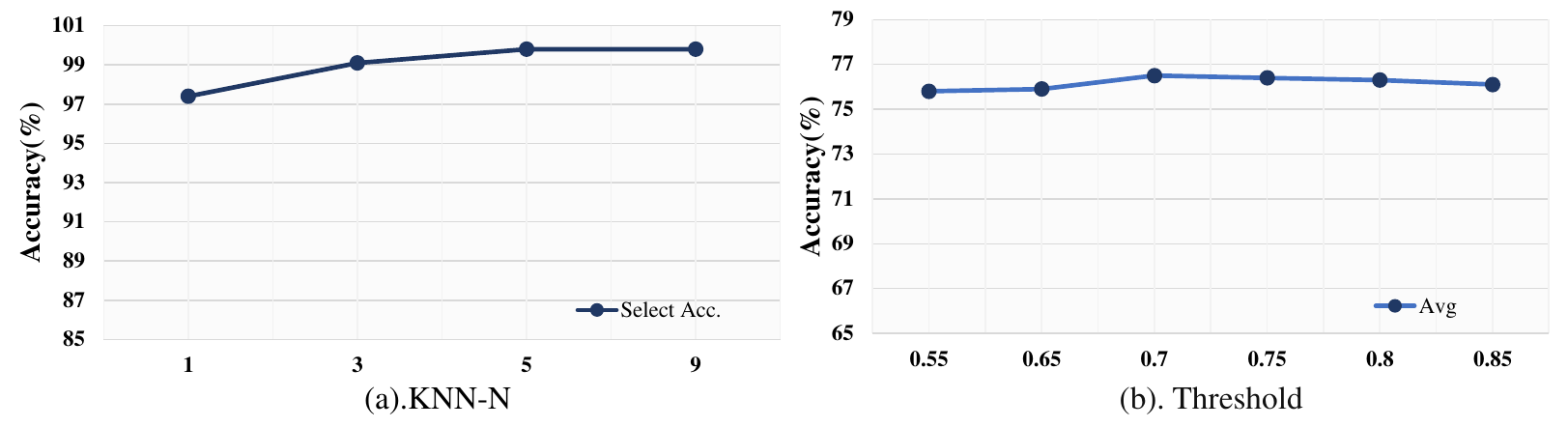}}
 \vspace{-3pt}
\caption{Impact of different Numbers of KNN-N and the threshold. The experiment is based on full-shot MTIL. If the distribution score of the test sample is below the threshold, it is classified as belonging to an unseen domain and processed by the AFFA module.}
\vspace{-20pt}
\label{fig:aba_select_thre}
\end{center}
\end{figure}

\noindent\textbf{Implementation Detail.}
All experiments were conducted using the PyTorch framework on an RTX 4090 24GB GPU. For structural details, we set the rank of both matrix \( A \) and matrix \( B \) to 16 for the experts in ABFA, and similarly, the adapter in AFFA is configured with a rank of 16~(Table~\ref{tab:supp_lora_rank}). For the domain distribution selector, the number of prototypes \( K \) for each dataset is set to 5 (see Fig.~\ref{fig:aba_select_thre} (a)), and the threshold for determining an unseen domain is set to 0.75~ (see Fig.~\ref{fig:aba_select_thre} (b)). Regarding training details, the full-shot MTIL experiments employ a learning rate of \( 1 \times 10^{-3} \) with a batch size of 16, while the few-shot MTIL experiments utilize a learning rate of \( 5 \times 10^{-4} \) with a batch size of 32.

\noindent\textbf{Compare methods.} In the MTIL and MTIL-FS tasks, we follow the ZSCL~\cite{zscl} method and compare it with traditional incremental learning methods such as LwF~\cite{LWF}, LwF-VR~\cite{lwm}, and WiSE-FT~\cite{wise}. Additionally, we follow the DIKI~\cite{zscl-prompt} method and compare it with parameter-tuning incremental learning approaches, including L2P~\cite{l2p}, Dual-prompt~\cite{dualprompt}, and S-prompt~\cite{S-prompts}. Furthermore, we compare our method with representative MTIL approaches such as ZSCL~\cite{zscl}, DIKI~\cite{zscl-prompt}, and BCL~\cite{zscl-moe}. To ensure a fair comparison, all methods use the same backbone (CLIP with ViT-B/16).

For the CIL task, we follow the ZSCL~\cite{zscl} method and compare it with traditional incremental learning methods such as LwF~\cite{LWF}, iCaRL~\cite{ICARL}, and LwF-VR~\cite{lwm}, as well as with MTIL approaches ZSCL~\cite{zscl} and BCL~\cite{zscl-moe}.

For domain incremental learning (DIL), we follow the S-prompts method and compare it with other ViT-B/16-based incremental learning methods, including DyTox~\cite{dytox}, LwF~\cite{LWF}, L2P~\cite{l2p}, and Dual-prompt~\cite{dualprompt}. Additionally, to ensure fairness, we compare our method with the latest DIL methods, S-prompts~\cite{S-prompts} and PINA~\cite{wang2024non}, as well as the MTIL method BCL~\cite{zscl-moe}, using the same model (CLIP with ViT-B/16).

Notably, the DIKI~\cite{zscl-prompt} method exhibits poor performance on both CIL and DIL tasks and is therefore excluded from our experiments. Moreover, since the ZSCL~\cite{zscl} method requires task IDs during inference, it is not applicable to DIL tasks and is thus omitted from our DIL experiments.

\section{Task Selection Accuracy of DDS}
\label{sec:task_select}
Table \ref{tab:task_select} presents the selection accuracy of our domain distribution selector~(DDS) on previously learned tasks. It can be observed that with DDS, we are able to accurately assign each test sample to its corresponding task during the testing phase, thereby effectively preventing catastrophic forgetting of knowledge specific to each task.
\begin{table*}[t]
\setlength\tabcolsep{5pt}
\centering
\label{tab:assignment_accuracy}
\fontsize{8pt}{10pt}\selectfont
\resizebox{0.9\textwidth}{!}{
\begin{tabular}{@{}lccccccccccc@{}}
\toprule
& \rot{Aircraft} & \rot{Caltech101} & \rot{CIFAR100} & \rot{DTD} & \rot{EuroSAT} & \rot{Flowers} & \rot{Food} & \rot{MNIST} & \rot{OxfordPet} & \rot{Cars} & \rot{SUN397} \\ \midrule
Aircraft & 100.0 & \textbf{-} & \textbf{-} & \textbf{-} & \textbf{-} & \textbf{-} & \textbf{-} & \textbf{-} & \textbf{-} & \textbf{-} & \textbf{-} \\
Caltech101 & 100.0 & 99.8 & \textbf{-} & \textbf{-} & \textbf{-} & \textbf{-} & \textbf{-} & \textbf{-} & \textbf{-} & \textbf{-} & \textbf{-} \\
CIFAR100 & 100.0 & 99.8 & 100.0 & \textbf{-} & \textbf{-} & \textbf{-} & \textbf{-} & \textbf{-} & \textbf{-} & \textbf{-} & \textbf{-} \\
DTD & 100.0 & 99.7 & 100.0 & 99.7 & \textbf{-} & \textbf{-} & \textbf{-} & \textbf{-} & \textbf{-} & \textbf{-} & \textbf{-} \\
EuroSAT & 99.8 & 99.7 & 100.0 & 99.7 & 100.0 & \textbf{-} & \textbf{-} & \textbf{-} & \textbf{-} & \textbf{-} & \textbf{-} \\
Flowers & 99.8 & 99.6 & 100.0 & 99.7 & 100.0 & 100.0 & \textbf{-} & \textbf{-} & \textbf{-} & \textbf{-} & \textbf{-} \\
Food & 99.8 & 99.8 & 100.0 & 99.7 & 100.0 & 99.9 & 99.6 & \textbf{-} & \textbf{-} & \textbf{-} & \textbf{-} \\
MNIST & 99.8 & 99.8 & 100.0 & 99.7  & 99.7 & 99.9 & 99.6 & 100.0 & \textbf{-} & \textbf{-} & \textbf{-} \\
OxfordPet & 99.8 & 99.7 & 100.0 & 99.7 & 99.6 & 99.9 & 99.7 & 100.0 & 100.0 & \textbf{-} & \textbf{-} \\
Cars & 99.8 & 99.8 & 100.0 & 99.7 & 99.6 & 99.8 & 99.8 & 100.0 & 100.0 & 100.0 & \textbf{-} \\
SUN397 & 99.8& 	99.8& 	99.8	& 99.7& 	99.6& 	99.7	& 99.6	& 99.7	& 99.8& 	99.8	& 99.9\\
\bottomrule
\end{tabular}%
}
\caption{\textbf{Task selection accuracy (\%) for test data.} Each row represents the selection accuracy on each dataset of the model trained after the corresponding task.}
\label{tab:task_select}
\end{table*}

\section{Additional Ablation Experiments}
\label{sec:add_ablation}

\subsection{Study on Application Strategies of AFFA}
\label{sec:add_AFFA}

\begin{table*}[!h]
    \centering
    \resizebox{\linewidth}{!}{
    \begin{tabular}{l c ccccccccccc}
    \toprule
    \fontsize{13pt}{9pt}\selectfont
    & \rotatebox{45}{Strategy} & \rotatebox{45}{Caltech101} & \rotatebox{45}{CIFAR100} & \rotatebox{45}{DTD} & \rotatebox{45}{EuroSAT} & \rotatebox{45}{Flowers} & \rotatebox{45}{Food} & \rotatebox{45}{MNIST} & \rotatebox{45}{OxfordPet} & \rotatebox{45}{Cars} & \rotatebox{45}{SUN397} & \rotatebox{45}{Average} \\
    \midrule
    \textbf{Transfer} \\
    \quad Ours & t & 88.4 & 68.2 & 44.7 & 55.3 & 71.0 & 88.5 & 59.5 & 89.0 & 64.7 & 65.0 & 69.4 \\
    \quad Ours & v & 87.2 & 67.7 & 45.3 & 53.9 & 71.0 & 88.0 & 63.5 & 89.0 & 63.7 & 66.9 & 69.6 \\
    \quad Ours & t/v & 88.6 & 67.9 & 45.7 & 54.8 & 71.3 & 88.5 & 64.4 & 89.6 & 64.7 & 67.0 & 70.3 \\
    \bottomrule
    \end{tabular}}
        \caption{\textbf{Impact of Adding AFFA to Different Branches on Transfer Performance.} We find that adding adapters to both the visual and textual encoders of CLIP yields the most significant improvement in transfer performance.}
    \label{tab:branch_strategy}
\end{table*}
To appropriately apply the AFFA module, we experimented with three strategies: applying it only to the CLIP text branch (the first row of Table~\ref{tab:branch_strategy}), applying it only to the CLIP visual branch (the second row of Table~\ref{tab:branch_strategy}), and applying it to both the CLIP visual and text branches simultaneously (the third row of Table~\ref{tab:branch_strategy}). We found that the most significant improvement in transferability was achieved when AFFA was applied to both the visual and text branches. We attribute this to the simultaneous adjustment of the visual and textual feature spaces, which facilitates better alignment between the visual and textual modalities.

\subsection{Ablation Study on the Hyperparameters of AFFA}
\label{sec:add_hy_affa}
\begin{table*}[!h]
    \centering
    \resizebox{\linewidth}{!}{
    \begin{tabular}{l c ccccccccccc}
    \toprule
    \fontsize{13pt}{9pt}\selectfont
    & \rotatebox{45}{Rank} & \rotatebox{45}{Caltech101} & \rotatebox{45}{CIFAR100} & \rotatebox{45}{DTD} & \rotatebox{45}{EuroSAT} & \rotatebox{45}{Flowers} & \rotatebox{45}{Food} & \rotatebox{45}{MNIST} & \rotatebox{45}{OxfordPet} & \rotatebox{45}{Cars} & \rotatebox{45}{SUN397} & \rotatebox{45}{Average} \\
    \midrule
    \textbf{Transfer} \\
    \quad Ours & 1 & 88.4 & 67.7 & 45.6 & 53.3 & 70.9 & 88.5 & 63.7 & 90.0 & 64.7 & 66.7 & 69.9 \\
    \quad Ours & 2 & 88.5 & 67.7 & 45.6 & 53.5 & 71.0 & 88.6 & 62.8 & 90.0 & 64.7 & 66.7 & 69.9 \\
    \quad Ours & 4 & 88.5 & 67.7 & 45.6 & 53.7 & 71.0 & 88.5 & 62.0 & 89.9 & 64.5 & 66.9 & 69.8 \\
    \quad Ours & 8 & 88.6 & 67.8 & 45.6 & 53.4 & 71.2 & 88.5 & 62.7 & 89.9 & 64.6 & 66.9 & 69.9 \\
    \quad Ours & 16 & 88.6 & 67.9 & 45.7 & 54.8 & 71.3 & 88.5 & 64.4 & 89.7 & 64.7 & 67.0 & 70.3 \\	
    \quad Ours & 32 & 88.5 & 67.9 & 45.7 & 53.9 & 71.1 & 88.6 & 59.9 & 89.6 & 64.6 & 66.9 & 69.7 \\
    \bottomrule
    \end{tabular}}
        \caption{\textbf{Impact of Different Ranks on the AFFA Capability.} We find that when the rank is set to 16, the domain-invariant module exhibits the strongest ability to extract domain-invariant information from each dataset, achieving the best generalization performance.}
    \label{tab:supp_lora_rank}
\end{table*}
Table~\ref{tab:supp_lora_rank} presents the impact of different rank values on the transferability of AFFA. Ultimately, we selected 16 as the rank of LoRA in the AFFA module.

\subsection{Additional Results of Multiple B}
\begin{table*}[!h]
    \centering
    \resizebox{\linewidth}{!}{
    \begin{tabular}{l cccccccccccc}
    \toprule
    \fontsize{13pt}{9pt}\selectfont
    & \rotatebox{45}{Aircraft} & \rotatebox{45}{Caltech101} & \rotatebox{45}{CIFAR100} & \rotatebox{45}{DTD} & \rotatebox{45}{EuroSAT} & \rotatebox{45}{Flowers} & \rotatebox{45}{Food} & \rotatebox{45}{MNIST} & \rotatebox{45}{OxfordPet} & \rotatebox{45}{Cars} & \rotatebox{45}{SUN397} & \rotatebox{45}{Average} \\
    \midrule
    \textbf{Last} \\
    \quad Ours~(1A/1B) & 32.1& 90.3 & 	73.9 & 	61.5	 & 83.1	 & 90.9	 & 87.2 & 	89	 & 89.2	 & 69.4	 & 70.5 & 	76.1\\
   
    \quad Ours~(1A/4B) &  \textbf{44.7}  & 	\textbf{92.3}  & \textbf{75.1}  & 63.9  & \textbf{89.6}  & 	\textbf{93.5}  & \textbf{87.8}  & \textbf{91.5}  & \textbf{89.5}  & {71.7}  & \textbf{73.7} &  \textbf{79.4}  \\
    \bottomrule
    \end{tabular}}
        \caption{Ablation studies on incremental Mixture-of-Experts
Adapters (ABFA). “1A/nB” indicates that each expert has one low-
rank matrix A and N high-rank matrices B, respectively.}
    \label{tab:supp_1A/nB}
\end{table*}
As shown in Table \ref{tab:aba_ima} in the main paper and Table \ref{tab:supp_1A/nB}, we observed that, compared to the original structure (1A/1B), using multiple B structures consistently enhanced the last accuracy across tasks. Notably, the best-performing 1A/4B structure yielded only a \textbf{1.1\%} improvement on the MTIL task but a substantial \textbf{3.1\%} boost on the MTIL-FS task, demonstrating the effectiveness of the multi-head architecture in few-shot learning. The significant improvements on fine-grained datasets (e.g., aircraft, dtd) and more complex datasets (e.g., sun397) further validate the method's ability to capture task-specific details and significantly enhance few-shot learning capabilities.

\subsection{Additional Hyper-parameter Sensitivity Analysis}
\label{sec:add_hyper}

\noindent\textbf{Hyper-parameter Sensitivity Analysis:}
\begin{itemize}
    \item \textbf{Impact of KNN-N.} As shown in Fig.~\ref{fig:aba_select_thre} (a), we investigated the impact of the number $N$ in KNN on the accuracy of task ID selection. The experiments were conducted on an MTIL task featuring T task-specific routers. The results indicate that as $N$ increases, the selection accuracy improves, particularly when $N$ is small. When $N$ exceeds 5, the selection accuracy tends to stabilize. Therefore, we uniformly adopt $N=5$ in our experiments.
     \item \textbf{Hyper-parameter Threshold.} In Fig.~\ref{fig:aba_select_thre} (b), we investigate the impact of the distribution score threshold on model performance. The experiments were conducted on an MTIL task featuring $T$ task-specific routers. The results indicate that our model exhibits high robustness to the distribution score threshold, achieving the best performance within the range of 0.7 to 0.8. Ultimately, we select 0.75 as the threshold for the distribution score.
\end{itemize}


\end{document}